\documentclass[conference]{IEEEtran}
\usepackage{amsmath,amssymb,amsfonts}
\usepackage{algorithmic}
\usepackage{subcaption}
\usepackage{graphicx}
\usepackage{textcomp}
\usepackage{xcolor}
\usepackage{blindtext}
\usepackage{lastpage}
\usepackage{fancyhdr}
\usepackage{adjustbox}
\usepackage{booktabs}
\usepackage[
    backend=biber,
    style=ieee,
]{biblatex}
\def\BibTeX{{\rm B\kern-.05em{\sc i\kern-.025em b}\kern-.08em
    T\kern-.1667em\lower.7ex\hbox{E}\kern-.125emX}}
\begin{document}

\title{Using CSNNs to Perform Event-based Data Processing \& Classification on ASL-DVS}

\makeatletter
\newcommand{\linebreakand}{%
  \end{@IEEEauthorhalign}
  \hfill\mbox{}\par
  \mbox{}\hfill\begin{@IEEEauthorhalign}
}
\makeatother

\author{
    \IEEEauthorblockN{
        Ria Patel,
        Sujit Tripathy, Zachary Sublett, Seoyoung An and Riya Patel}
    \IEEEauthorblockA{
        \textit{
            Department of Electrical Engineering \& Computer Science,
        }
    }
    \IEEEauthorblockA{
        \textit{
            University of Tennessee, Knoxville, TN 37916, USA
        }
   }
}
\maketitle

\maketitle
\begin{abstract}

Recent advancements in bio-inspired visual sensing and neuromorphic computing have led to the development of various highly efficient bio-inspired solutions with real-world applications. One notable application integrates event-based cameras with spiking neural networks (SNNs) to process event-based sequences that are asynchronous and sparse, making them difficult to handle. In this project, we develop a convolutional spiking neural network (CSNN) architecture that leverages convolutional operations and recurrent properties of a spiking neuron to learn the spatial and temporal relations in the ASL-DVS gesture dataset. The ASL-DVS gesture dataset is a neuromorphic dataset containing hand gestures when displaying 24 letters (A to Y, excluding J and Z due to the nature of their symbols) from the American Sign Language (ASL). We performed classification on a pre-processed subset of the full ASL-DVS dataset to identify letter signs and achieved 100\% training accuracy. Specifically, this was achieved by training in the Google Cloud compute platform while using a learning rate of 0.0005, batch size of 25 (total of 20 batches), 200 iterations, and 10 epochs.

\end{abstract}

\section{Introduction}
The application of deep learning has been instrumental in solving numerous computer vision problems, notably image classification~\cite{b1}, object detection and semantic segmentation~\cite{b2}, and gesture recognition~\cite{b3,b4,b5,b6}, etc. For some specific applications such as static image recognition, Deep Neural Networks (DNNs) have been demonstrated to outperform human performance~\cite{b1,b7}. Training state-of-the-art deep learning models is computationally expensive, leading to high energy usage. To reduce the energy cost of future AI systems, researchers are exploring the use of energy-efficient, brain-inspired spiking neural networks (SNNs) in various fields~\cite{b8}. SNNs typically take inputs from neuromorphic sensors that generate spikes with changes in input signals. Next, SNNs interpret such spikes, which contain sparsely encoded spatial and temporal information. SNNs implemented on neuromorphic hardware allow the hardware to utilize sparse activations within the SNNs to gain significant power and latency improvements compared to conventional hardware performing the same tasks~\cite{b8}.

For data acquisition, dynamic vision sensors (DVS) are used to capture individual pixel changes within each frame of a recording, making them capable of capturing motion very similar to humans. Pixels that change are considered events whereas those that don’t change are not considered events by the camera. Next, the event-based data is encoded into spikes such that the SNN can interpret the spatiotemporal features.

Conventional frame-based cameras capture information from all pixels within the frame, making the image densely rich in information and memory demanding. Event-based cameras mitigate this issue by reducing this pixel information to make them sparsely encoded. As a result, these DVS cameras consume low power enabling operation at a finer time resolution of a few milliseconds~\cite{b8}. 

Recently, SNNs have been used for problems that involve data with strong spatiotemporal correlations such as hand gesture recognition, object tracking, action and object recognition etc.~\cite{b9, b10, b11}. In this work, we contribute a CSNN that performs classification on event-based data from ASL-DVS dataset ~\cite{b13}. The rest of the report is organized as follows: Section II includes a
literature review that provides some background on CSNN architecture as well as discusses some prior work that utilized SNNs including CSNNs. Section III presents the proposed CSNN framework for ASL-DVS data processing and classification. Section IV discusses the characteristics of the ASL-DVS dataset and its preprocessing steps in detail. Next, we present the experimental set-up and results before concluding the report.

\section{Literature Review}

In this section, we first present the SNN and CSNN model architecture. Next, we discuss some prior works that utilized SNNs including CSNNs for event-based data processing.

\subsection{SNNs}
An SNN shares similarities with an artificial neural network (ANN). Both models calculate neuron outputs using the weighted sums of inputs. However, SNNs process single-bit spikes as input, which are sparse in both space and time. Also, SNNs employ different types of activations compared to ANNs. The weights contribute to the membrane potential, and when this potential reaches a certain threshold, the neuron generates a spike. Spikes introduce a temporal component to the data processed by an SNN. In the ASL-DVS dataset, this temporal aspect is measured in $\mu$s to capture subtle movements occurring within the images ~\cite{b8}.

There are several different types of spiking neuron models, but one of the most commonly used is called the Leaky Integrate and Fire (LIF) neuron. This neuron is closely modeled after biological neurons, making it suitable for neural networks. It integrates the weights of its input over time until it reaches or exceeds a certain threshold and then, the neuron will fire a spike. The membrane potential will decay towards a lower value at each step controlled by the decay rate $\beta$, enforcing the "leaky" aspect of the neuron. Figure~\ref{lif_neuron} demonstrates this feature, which is taken from ~\cite{b8}.

\begin{figure}
    \centering
    \includegraphics[width=0.5\textwidth]{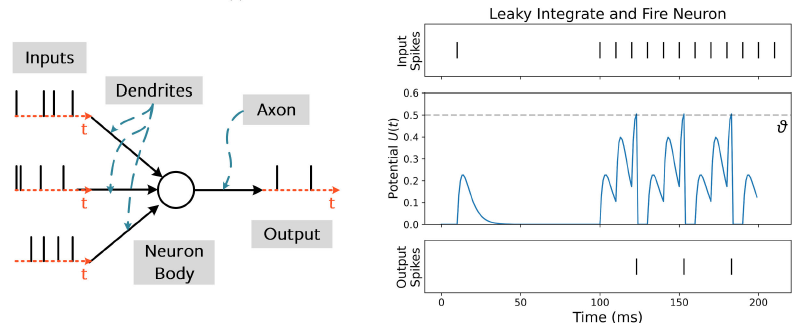}
    \caption{LIF operation in SNN ~\cite{b8}}
    \label{lif_neuron}
\end{figure}

The LIF membrane is modeled with:
\begin{equation}
    \tau \frac{dU(t)}{dt} = -U(t) + RI_{in}t
\end{equation}

This equation models how the membrane potential changes over time. Where $I_{in}$ represents the input-weight matrix, $R$ represents the membrane resistance, and the membrane potential at a given time t is represented by $U(t) = \beta U(t - 1) + (1 - \beta) I_{in}(t)$, where $\beta$ is the decay rate of the membrane potential.

\subsection{CSNNs}

A convolutional spiking neural network typically comprises spiking convolution operations and spiking encoding. Assume an input sequence of $S(n)$, for $n = 0, 1, 2,..., S(n)$ is a tensor at every time step that has shape ${u,v}$, where u and v are the width and height of each frame. Output from event cameras can be processed using a variable batch window to achieve the desired temporal resolution. For example, a 3 s $ 240 \times 180$ resolution events data stream with 100 ms temporal resolution will generate $S(n), n = 0, 1, 2, ..., 30$. For each segment, the tensor shape is (240,180). Next, the input tensor is convolved with a 2-D convolutional kernel to generate a spiking feature map ~\cite{b14, b15, b16}. The convolution operation can be described using the following equations ~\cite{b16}:

\begin{equation}
    y[i,j] = \sum_{j = -\infty}^{\infty} \sum_{i = -\infty}^{\infty} = x(i + m, j + n).K(m,n)
\end{equation}

Where, $y[i,j]$ represents the spiking feature map, $x$ is the input tensor, $K$ is the kernel, $(i,j)$ are coordinates on the feature map and $(m,n)$ are the coordinates in the kernel. Neurons in the feature map receive spikes, which are aggregated to generate membrane potential. 

The convolutional layers extract features. The LIF neurons at the output of each layer then gather all the input spikes and output a spike when the membrane potential crosses the set threshold. Lastly, similar to the traditional convolutional neural networks (CNNs), the last layer of CSNNs is a fully connected (FC) layer with LIF neurons and has an output shape equal to the number of classes for classification.

\subsection{Prior Technical Work}
In this subsection, we review prior works discussing applications of SNNs and CSNNs for event-based data processing.

Amir et al. implemented a gesture recognition system that utilized a TrueNorth neurosynaptic processor alongside a DVS for real-time hand gesture recognition with minimal power consumption. They achieved an out-of-sample accuracy of approximately 96.5\% on a diverse dataset, demonstrating the robustness of the approach for various hand gestures under different lighting conditions~\cite{b9}.

Weijie et. al used CSNN for EMG (Electromyogram)-based hand gesture recognition. They utilized the Strathclyde and CapgMyo open-source datasets for model development. The average recognition accuracy was observed to be around 98.76\% based on the Strathclyde dataset and 98.21\% based on the CapgMyo open source dataset~\cite{b15}.

Xing et al. introduced a spiking convolutional recurrent neural network (SCRNN) model for event-based hand gesture recognition, using the DVS gesture dataset. They achieved an accuracy of 96.59\% for 10-class gesture recognition and 90.28\% for 11-class gesture recognition~\cite{b14}.

Samadzade et al. investigated the spatiotemporal feature extraction capabilities of CSNNs for event-based data processing. They observed superior performance of CSNNs methods on N-MNIST (97.6\%), DVS-CIFAR10 (69.2\%), and DVS-Gesture (96.7\%) datasets. Additionally, the model achieved results comparable to ANN methods on UCF-101 (42.1\%) and HMDB-51 (21.5\%) datasets~\cite{b17}.

The availability of comprehensive neuromorphic datasets plays a crucial role in the increased implementation of SNN-based architecture. Yin et. al~\cite{b13} and Orchard et. al~\cite{b12} created neuromorphic ASL-DVS and MNIST datasets. Eshraghian et al. compiled a comprehensive list of neuromorphic vision and audio benchmark datasets, widely accessed for SNN-based model development~\cite{b8}. Moreover, efforts have been made to convert traditional computer vision datasets to neuromorphic vision datasets. A method for converting static image datasets into spiking neuromorphic datasets using an actuated pan-tilt camera platform was presented. By simulating motion updates through sensor movement, this approach enabled the utilization of popular image datasets like MNIST and Caltech101 for spike-based recognition algorithms~\cite{b12}.

Lastly, to bridge the gap between deep learning and SNNs, Eshraghian et al. presented principles from deep learning, gradient descent, backpropagation, and neuroscience to develop SNNs. By elucidating the relationship between encoding data as spikes and the learning process, this work helps to advance the understanding and application of SNNs in NN research and development~\cite{b8}.

\section{Technical Approach}
 
In this section, we present the overall deep learning pipeline for ASL-DVS data processing and classification. Then, we briefly discuss data selection and processing, the architecture of the CSNN model trained for ASL-DVS classification, and the CSNN learning method.

\subsection{Deep Learning Pipeline}
Figure~\ref{pipeline} shows the overall deep-learning pipeline. We start with acquiring neuromorphic ASL-DVS datasets. Subsequently, we pre-process the data to bring it to a format so that it can be imported into the CSNN model. Next, we have an iterative process of model training and tuning to improve the CSNN model performance. Then, we validate our model that performs ASL-DVS gesture classification.

\subsection{Data Selection and Pre-Processing}
For this project, we have used the ASL-DVS dataset to perform event-based data processing and classification. ASL-DVS dataset was acquired in AEDAT format and then, IniVation DV Processing software was used to pre-process it before the data could be imported into a CSNN model for training and validation. In the next section, we discuss the characteristics of the ASL-DVS dataset and its preprocessing steps in detail.

\subsection{Network Architecture}
The CSNN model used to process the neuromorphic ASL-DVS dataset is built upon a similar CSNN implemented for N-MNIST data processing and classification ~\cite{b8}. However, we updated the network
architecture to accommodate the ASL-DVS dataset. The model architecture comprises of three convolutional layers and a FC layer with 24 outputs for the 24 classes of the ASL-DVS dataset. Figure \ref{Net} is a diagram showing the model's architecture. Each convolutional layer includes 5x5 filters with layers 1, 2, and 3 containing 12, 32, and 45 filters respectively. Each convolutional layer outputs to a 2x2 max pool layer and then to a LIF (leaky integrate and fire) neuron. The model then feeds into a FC layer with 24 output LIF neurons, that will determine the class or letter in the ASL-DVS dataset that the input belongs to. Figure \ref{Net1} shows the details involved in calculating the size of CSNN layers.

\subsection{CSNN Learning Method}
For model training, the Adam optimizer is utilized. The loss function employed is the Mean Square Error (MSE) of the spike count, calculating the MSE between the correct and incorrect spike count estimates. These estimates are typically expressed as proportions of spikes over the total number of time steps. The target proportions for the loss function can vary, but a common choice is an 80\% correct rate and a 20\% incorrect rate~\cite{b8}, which is used in this model.

To perform backpropagation using spikes, a specific method known as BackPropagation Through Time (BPTT) is utilized. This algorithm computes gradients by determining the loss of all descendants in the computational graph and aggregating them. Consequently, the derivative of the loss with respect to the weights $\frac{\partial \mathcal{L}}{\partial W}$ is computed for each time step. The equation below illustrates how backpropagation computes this derivative ~\cite{b8}:

\begin{equation}
    \frac{\partial \mathcal{L}}{\partial W} = \sum_{t}\frac{\partial \mathcal{L}(t)}{\partial W} = \sum_{t}\sum_{t\le s}\frac{\partial \mathcal{L}(t)}{\partial W(s)}\frac{\partial \mathcal{W}(s)}{\partial W}
\end{equation}

This allows for the model to learn the parameters, but there is still the problem of calculating the gradients when spikes themselves are non-differentiable (also, called "dead neuron problem"). The adopted solution to this is known as the surrogate gradient method. Essentially, this swaps the non-differentiable function for one that is differentiable during the calculation of gradients.

\begin{figure}
    \centering
    \includegraphics[width=0.5\textwidth]{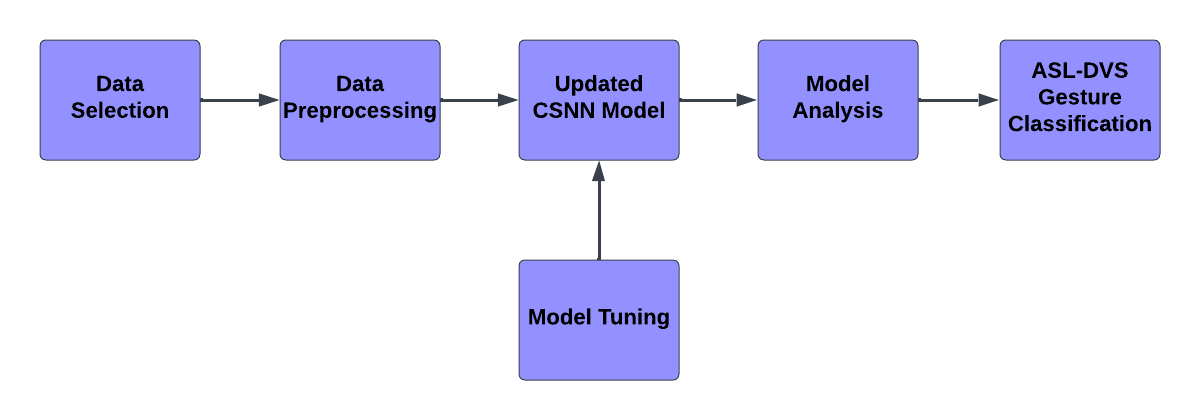}
    \caption{Pipeline of our approach}
    \label{pipeline}
\end{figure}

\begin{figure}
     \centering
     \includegraphics[width=1\linewidth]{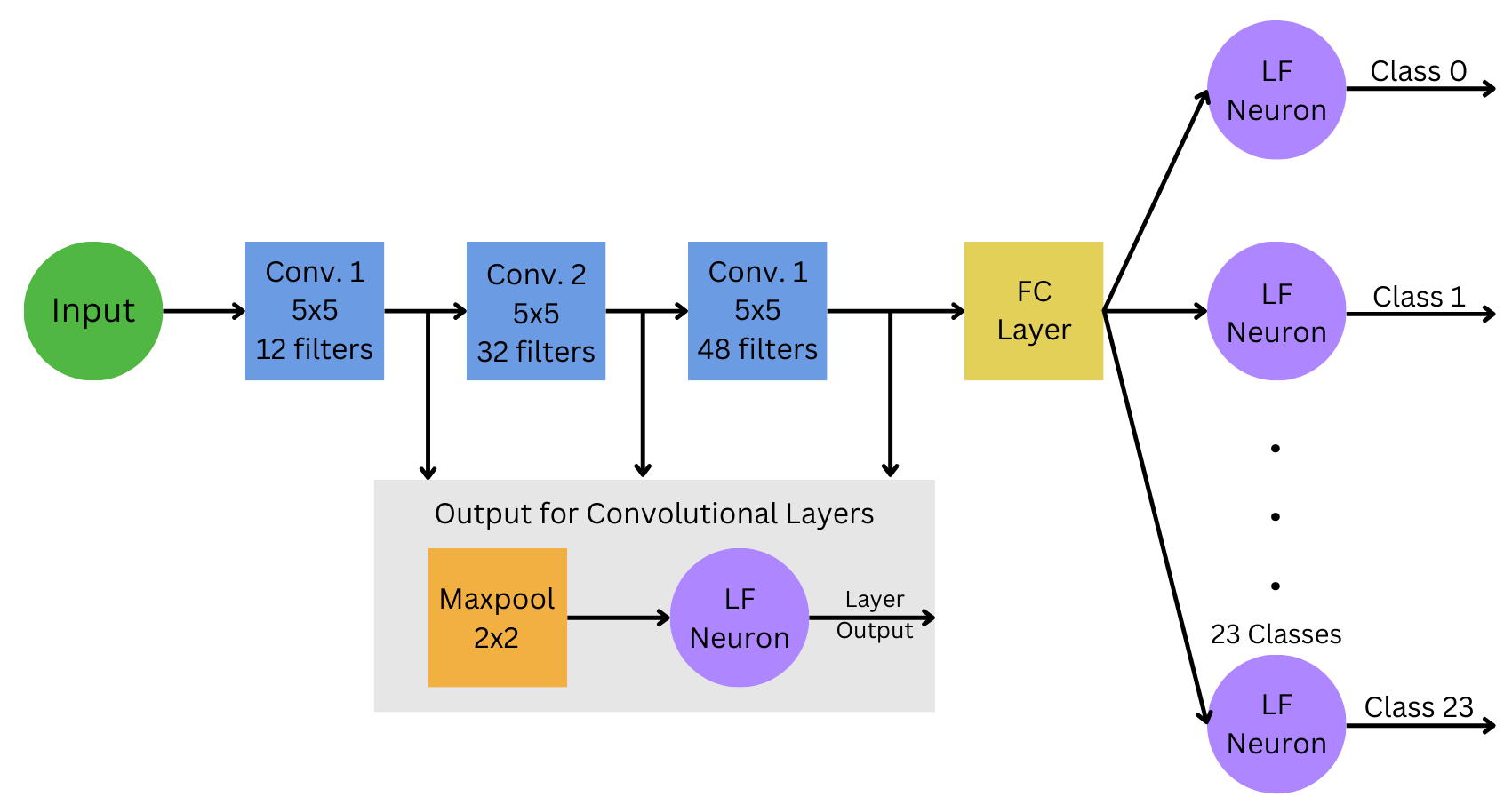}
     \caption{Model of CSNN Network Architecture}
     \label{Net}
 \end{figure}

\begin{figure}
    \centering
    \includegraphics[width=1\linewidth]{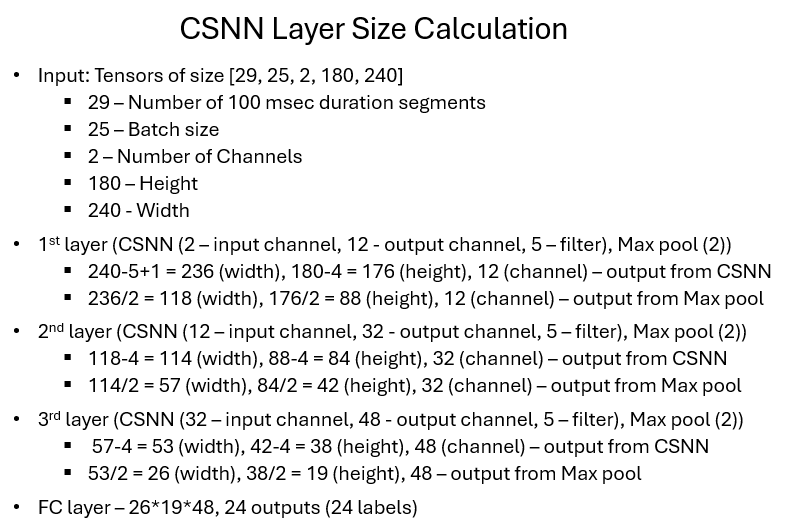}
    \caption{Calculation of Size of CSNN Layers}
    \label{Net1}
\end{figure}

\section{Dataset \& Implementation}
\subsection{Dataset Characteristics}

\begin{figure}
     \centering
     \includegraphics[width=\linewidth]{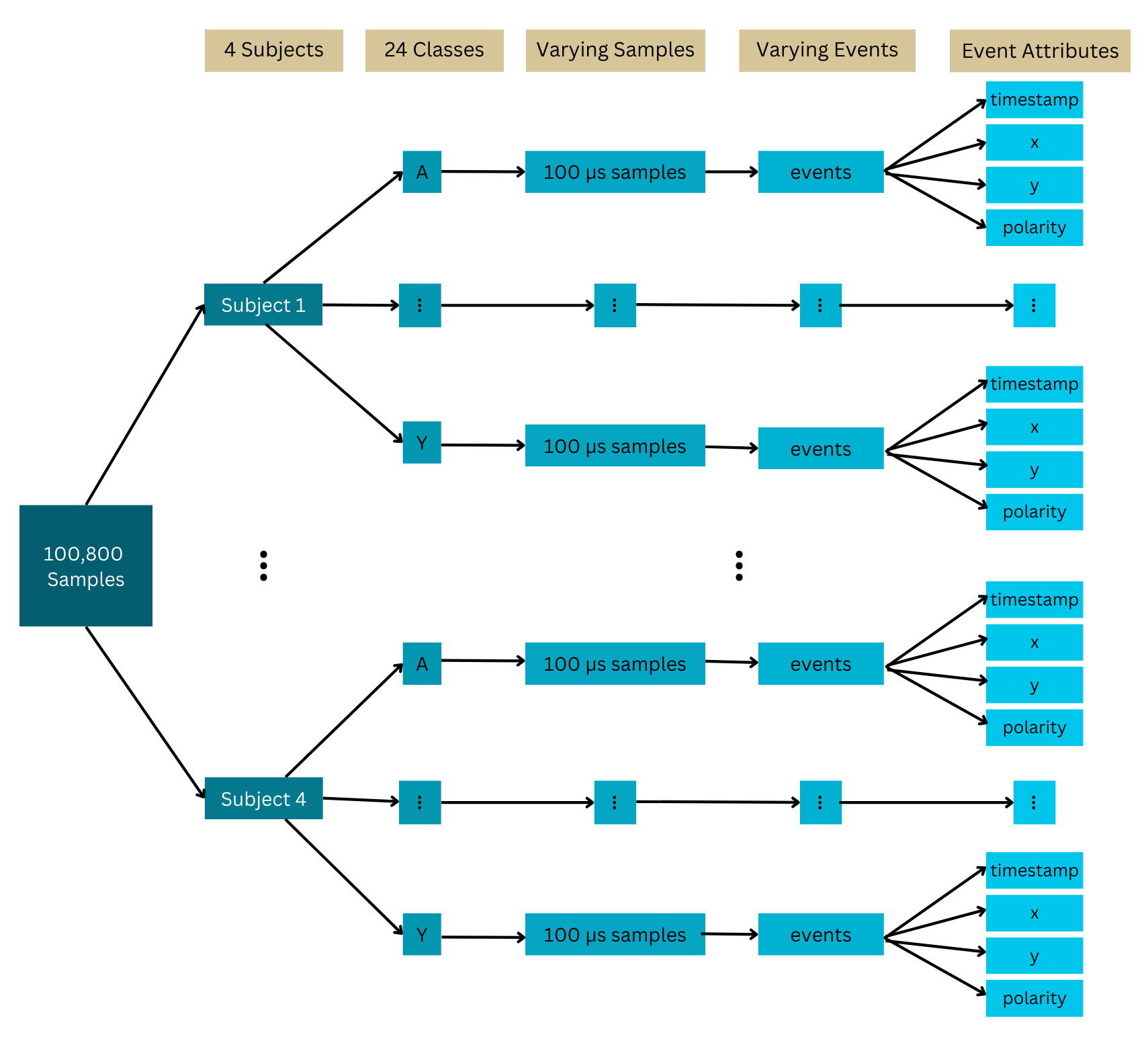}
     \caption{ASL-DVS dataset components}
     \label{components}
 \end{figure}

The ASL-DVS dataset was created using the IniVation DAVIS240C NVS event-based camera under constant lighting conditions, which captures at a $240 \times 180$ resolution and has a 12 µs latency~\cite{b22}. The events are captured and stored in the AEDAT 2.0 format, but a conversion process is used to achieve AEDAT 4.0, the newest AEDAT format that allows for the usage of the IniVation DV Processing software framework. Using this software, the AEDAT files were converted to CSV files to better view the event-based information and perform data exploration. 

\begin{figure}
    \centering
    \includegraphics[width=\linewidth]{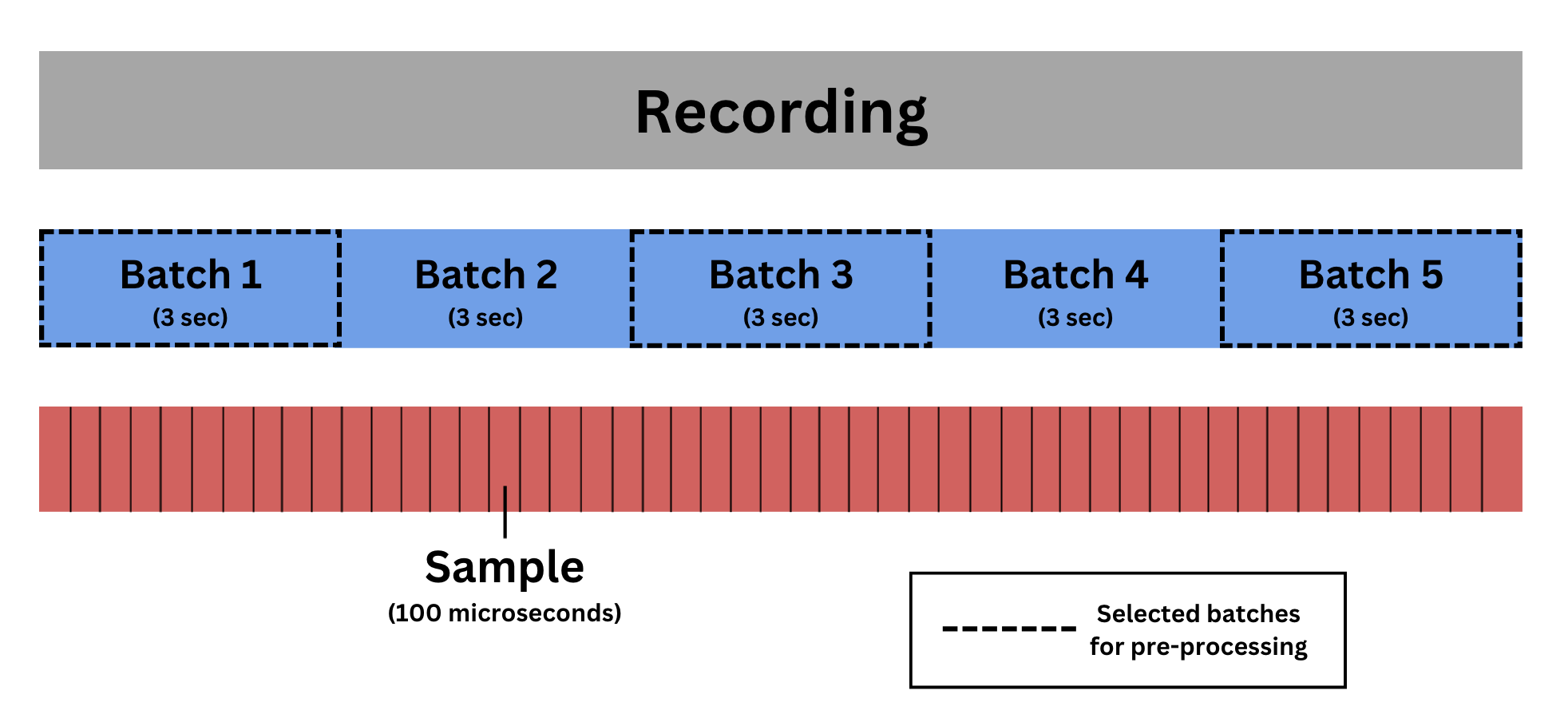}
    \caption{Dataset pre-process breakdown}
    \label{dataset_preprocess}
\end{figure}

This dataset contains recording data from five subjects who signed each letter of the alphabet, excluding J and Z. The ASL for J and Z are excluded from the dataset due to the nature of how those letters are signed (they include hand motion along with gesture). In this project, recording data from the first four subjects is used to train the CSNN. 

Figure~\ref{components} illustrates the components of the ASL-DVS dataset. The dataset comprises 100,800 samples, also referred to as "frames," each lasting approximately 100 microseconds ($\mu$s). The number of samples varies for each letter in each subject, and each sample contains a varying number of events detected by the DVS camera. The terminology used in the data pre-processing section is explained in Figure~\ref{dataset_preprocess}.

Each event has four attributes: timestamp, x, y, and polarity. The timestamp denotes the exact $\mu$s that this event took place in the specified pixel via x and y coordinates. Lastly, polarity denotes whether there is an increase or decrease in the pixel brightness or intensity, eventually to be encoded as a spike.

\subsection{Dataset Exploration}

To visualize the ASL-DVS dataset, we used the DV-Processing software to load each subject and each letter in the AEDAT4.0 format. The Python module library contains classes that load in each recording as a batch of events, which renders an image as shown in Figure~\ref{asl_f}. This image contains 1,515 events, creating the silhouette of Subject 1's hand signing the letter 'f' as they move it in an oscillating left-to-right motion.
The DVS camera picks up all movements and stores them as the black pixels shown in Figure~\ref{asl_f}, however, in the AEDAT files, the polarity field denotes them as either 0 or 1. This signifies that at the specific pixel, either there was a decrease or increase in pixel brightness. For the sake of visualization and ease of use for the SNN model, we encode all events that occur as 1 (black pixels) while the rest are encoded as 0 (white space).

\begin{figure}[t!]
    \centering
    \centering
    \includegraphics[width=\linewidth]{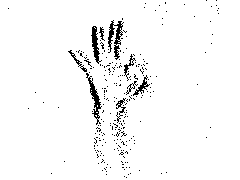}
    \caption{Subject 1 - Letter F}
    \label{asl_f}
\end{figure}

Figure~\ref{total_sample_distro} illustrates the overall sample distribution for each class across all subjects in the original ASL-DVS dataset. It is important to note that there are fewer samples for the letters Y and Z. This discrepancy is due to Subject 5 excluding recordings for the letters J and Y, while Subjects 1-4 exclude recordings for the letters J and Z.

\begin{figure}
    \centering
    \includegraphics[width=\linewidth]{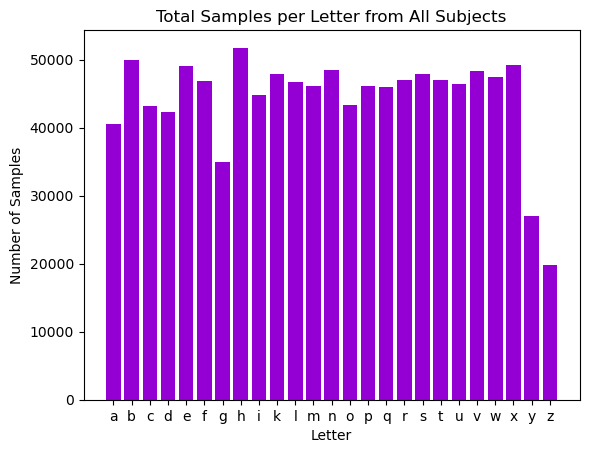}
    \caption{Original ASL-DVS sample distribution across all subjects}
    \label{total_sample_distro}
\end{figure}

Figure~\ref{vowel_distro} presents the sample distribution for each subject concerning vowel letters. Vowels were chosen for visualization as they are the most frequently used letters in English. Notably, Subject 5 contributes significantly to the sample distribution, indicating that Subject 5's dataset contains the longest recordings.

\begin{figure}
    \centering
    \includegraphics[width=\linewidth]{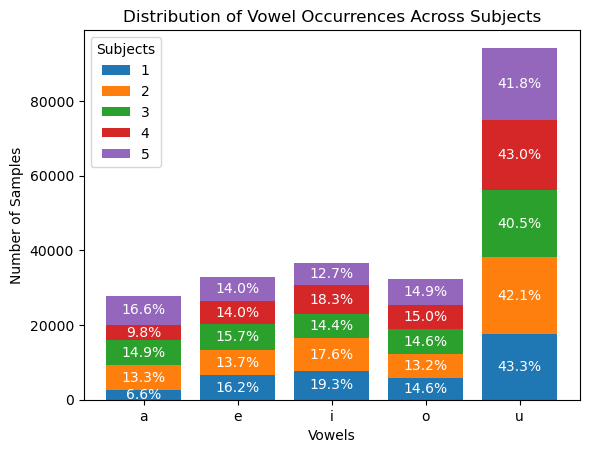}
    \caption{Sample distribution per subject for vowel letters}
    \label{vowel_distro}
\end{figure}

Figure~\ref{batch_samples} provides a visualization of the number of events captured from Subject 1's recording of the letter F. The recording is divided into 3-second batches, each containing a varying number of events within the time window. This figure illustrates the subject's hand movement patterns at the beginning, middle, and end of the recording, demonstrating how the oscillation path of the hand changes throughout the recording.

\begin{figure}
    \centering
    \includegraphics[width=\linewidth]{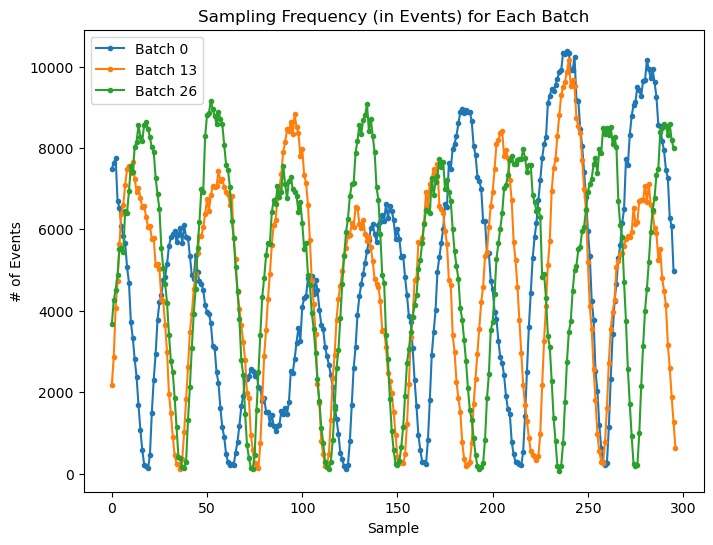}
    \caption{Events captured per sample in batches from different parts of Subject 1's Letter F recording.}
    \label{batch_samples}
\end{figure}

\subsection{Dataset Pre-processing}

We have extracted a subset from the ASL-DVS dataset by randomly choosing batches of events from each subject's recordings for a letter. Each recording varies in duration (e.g. 3 minutes, 5 minutes). We divide each recording into 3-second event batches, each containing a different number of events in each sample. Figure~\ref{asl_f} provides an example of an event batch, although it is not from the preprocessed dataset.

Following the event batch creation, we randomly select 3 event batches from each recording to include in our subset. For each subject and class, we save the randomly selected event batches into Numpy binary files. Subsequently, we combine all event batches from each subject and organize them into 24 classes.

To create the train and validate datasets, we randomly shuffle the event batches for each class. We then allocate 70\% of the event batches for training and reserve the remaining 30\% for validation. Our training set contains 495 event batches and our validation set contains 214 event batches.

\section{Experiments \& Results}

\subsection{Experimental Setup}
As shown in \ref{Net1}, the input to the model is a tensor of dimension: [time steps, batch size, channels, height, width]. An input event batch of 3 sec duration has 30 time steps with bin window duration of 100 milliseconds. The batch size is set at 25 implementing mini-batch gradient descent (GD). Also, the frame height and width are 180 and 240 respectively with 2 number of channels. The training dataset that consists of 495 event batches is passed onto the CSNN model and the model is trained based on the learning method discussed in section III. For this work, the 'Adam' optimizer is utilized with a learning rate = 0.0005, Betas = (0.9, 0.999). The loss function used is 'MSE of the spike count' with the 80\% correct rate and a 20\% incorrect rate. With 495 training data points (event batches) and a mini-batch batch size of 25, we have 20 iterations in each epoch. For training, we have experimented with varying numbers of epochs. Similarly, for validation, we have a batch size of 214 with all 214 event batches being processed simultaneously to generate validation results.

\subsection{Hardware}

Given the size of the dataset, we turned to Google Cloud Platform (GCP) for training our model using various hyperparameter combinations. When attempting to run the model locally on the dataset, we encountered issues with insufficient memory.

In terms of hardware, we employed the N2 machine series, featuring 8 Intel Cascade Lake and Ice Lake virtual CPUs (referred to as threads by GCP) and 32GB RAM. This series utilizes x86 architecture and the SCSI and NVMe disk interface type. We selected the N2 machine series due to its suitability for general-purpose workloads and batch processing, aligning perfectly with our requirements.

On GCP, we leveraged the Vertex AI Machine Learning platform to conduct our model training within a Jupyter Notebook. Before training, we set up a Bucket, a storage solution in GCP, to house our dataset. Subsequently, within the Vertex AI-hosted Jupyter Notebook, we trained the model using the dataset stored in the Bucket, which already contained the separated training and testing data.

\subsection{Results}
This subsection shows the results from our model for the best-performing hyper-parameters, in the form of loss and accuracy plots as well as a table.

Figures \ref{plot1} and \ref{plot10} show the model's training accuracy and loss plots for training cases with 1 epoch and 10 epochs each. This was the best training performance the model had achieved. It starts to stabilize around 90\%-100\% for the last several GD batches. Also, the loss is gradually decreasing across all batches. 

Figure \ref{valPlot} shows the validation loss and accuracy across models trained for 1, 3, 5, 9, and 10 epochs. The models that had been trained for longer epochs generally got a higher accuracy and a lower loss indicating that longer training will result in the model learning better and in turn performing better. 

Table \ref{results} shows an overview of results from the different epochs. As supported by the graphs, the training loss gradually decreases while the training accuracy increases with the number of epochs. It even hits 100\% accuracy for a single training batch (GD batch) as early as 5 epochs. Similarly, the validation loss also decreases while accuracy increases for each of the epochs. However, we also noted the mismatch between training and validation accuracy. This might indicate that the model was overfitting to the training dataset although it could otherwise be attributed to noise in event batches (some event batches captured background objects behind some subjects showing ASL gestures). This leads to the proposed future work, which may aim at exploring different regularization techniques to improve validation accuracy.
\begin{figure}
     \centering
     \includegraphics[width=0.95\linewidth]{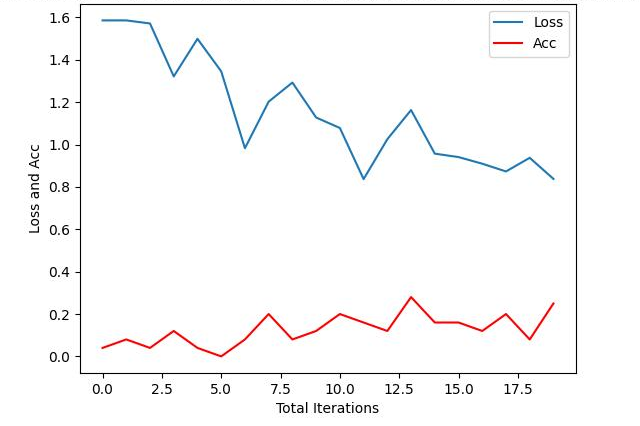}
     \caption{Training Accuracy and Loss plot for 1 Epoch}
     \label{plot1}
 \end{figure}

 \begin{figure}
     \centering
     \includegraphics[width=0.95\linewidth]{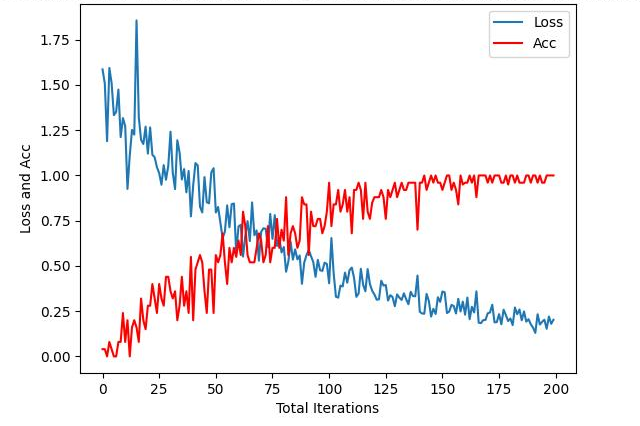}
     \caption{Training Accuracy and Loss plot for 10 Epochs}
     \label{plot10}
 \end{figure}

 \begin{figure}
     \centering
     \includegraphics[width=1\linewidth]{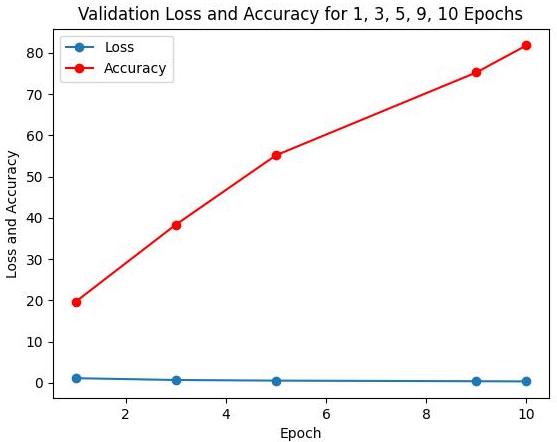}
     \caption{Validation Accuracy and Loss on Several Models}
     \label{valPlot}
 \end{figure}

\begin{table*}[htbp]
\begin{center}
\resizebox{0.75\textwidth}{!}{%
\begin{tabular}{@{}lllllllll@{}}
\toprule
\begin{tabular}[c]{@{}l@{}}No. of \\ Epochs\end{tabular} &
  \begin{tabular}[c]{@{}l@{}}No. of\\ Batches\end{tabular} &
  \begin{tabular}[c]{@{}l@{}}Batch \\ Size\end{tabular} &
  \begin{tabular}[c]{@{}l@{}}No. of \\ Iterations\end{tabular} &
  \begin{tabular}[c]{@{}l@{}}Learning \\ Rate\end{tabular} &
  \begin{tabular}[c]{@{}l@{}}Train\\ Loss\end{tabular} &
  \begin{tabular}[c]{@{}l@{}}Train \\ Acc(\%)\end{tabular} &
  \begin{tabular}[c]{@{}l@{}}Val\\ Loss\end{tabular} &
  \begin{tabular}[c]{@{}l@{}}Val\\ Acc(\%)\end{tabular} \\ \midrule
1  & 20 & 25 & 20  & 0.0005 & 0.85 & 26  & 1.12 & 20 \\
3  & 20 & 25 & 60  & 0.0005 & 0.62 & 63  & 0.68 & 38 \\
5  & 20 & 25 & 100 & 0.0005 & 0.35 & 100 & 0.53 & 55 \\
9  & 20 & 25 & 175 & 0.0005 & 0.21 & 100 & 0.38 & 75 \\
10 & 20 & 25 & 200 & 0.0005 & 0.19 & 100 & 0.34 & 81 \\ \bottomrule
\end{tabular}%
}
\end{center}

\caption{Experiment data for various hyperparameter combinations}
\label{results}
\end{table*}

\section{Conclusion}

For this project, we aimed to develop a CSNN model capable of processing and classifying the ASL-DVS dataset. To begin, we obtained the ASL-DVS dataset and pre-processed it to prepare it for training the CSNN model. Using GCP, we successfully developed and implemented the CSNN, achieving a high training accuracy of 100\% and a validation accuracy of 81\%. Our findings indicate that as the number of training epochs increased, the losses decreased and the accuracies improved. However, we observed a disparity between training and validation accuracies, suggesting the need for further research to explore different regularization techniques aimed at enhancing validation accuracies.

\section*{Acknowledgment}
We would like to express our gratitude to Dr. Charlie Rizzo from the TENNLab for his invaluable assistance in clarifying DVS datasets. Despite recently defending his Ph.D. thesis, he generously devoted his time to aid us in understanding the intricacies of neuromorphic dataset processing.

\printbibliography

\section{Appendices}

\subsection{Coding}
Our code repository~\cite{Patel_CSNN4ASLDVS} is organized into three main directories. The \texttt{animations/} directory contains visualization GIFs showcasing the event-based data. The \texttt{src/} directory houses our model code and dataloader scripts. The \texttt{utils/} directory contains code for data preprocessing, analysis, and visualization.

For our implementation, we utilize the PyTorch framework along with the snnTorch framework, which serves as a PyTorch extension for spiking neural networks. Additionally, we incorporate Tonic, a library that streamlines the loading of neuromorphic datasets into the PyTorch framework. Tonic enables the application of transforms specifically designed for spatiotemporal datasets, while also being compatible with native PyTorch transforms.

\subsection{Code Design}

\subsubsection{Data Pre-processing}
Our data pre-processing code utilizes the DV-Processing library mostly. In the `utils/` directory of our repository, we created a `helper\_funcs.py` that contains utility functions to perform batching of the recording data. These functions are used in the Jupyter notebooks `data\_explore.ipynb` and `data\_preprocess.ipynb`, which are to be run in the respective order. The `generate\_letter\_datasets.py` script, which must be run before the `data\_preprocess.ipynb` code, utilizes the other two scripts in the directory `export\_aedat4\_to\_csv.py` and `process\_aedat\_files.py` to create the training and validation datasets.

\subsubsection{Model}
The model design code was influenced by a snnTorch tutorial demonstrating the use of Convolutional Spiking Neural Networks (CSNNs) on the Neuromorphic MNIST (N-MNIST) dataset~\cite{b8, csnn_tutorial}. This tutorial provided insights into importing the DVS dataset into Tonic. 

To adapt the CSNN architecture for the ASL-DVS dataset, which has larger and non-square input dimensions compared to the $34 \times 34$ dimensions of the N-MNIST dataset, we made significant changes to the network structure. These changes included modifying the layers, adjusting the number of neurons in each layer, and refining the convolutional filters.

\end{document}